\pdfoutput=1

\documentclass[11pt]{article}

\usepackage[final]{coling}

\usepackage{times}
\usepackage{latexsym}

\usepackage[T1]{fontenc}

\usepackage[utf8]{inputenc}

\usepackage{microtype}

\usepackage{inconsolata}

\usepackage{graphicx}

\usepackage{latexsym}
\usepackage{amssymb}
\usepackage{amsmath}
\usepackage{amsthm}
\usepackage{booktabs}
\usepackage{enumitem}
\usepackage{graphicx}
\usepackage{color}
\usepackage{todonotes}
\usepackage{multirow}
\usepackage{multicol}
\usepackage{comment}

\title{Improving the Efficiency of Visually Augmented Language Models}

\author{Paula Ontalvilla \quad  Aitor Ormazabal \quad Gorka Azkune \\
        HiTZ Center - Ixa, University of the Basque Country (UPV/EHU) \\
        \texttt{\{paula.ontalvilla, aitor.ormazabal, gorka.azcune\}@ehu.eus}}

\begin{document}
\maketitle

\begin{abstract}
Despite the impressive performance of autoregressive Language Models (LM) %
it has been shown that due to reporting bias, LMs lack visual knowledge, i.e. they do not know much about the visual world and its properties. %
To augment LMs with visual knowledge, existing solutions often rely on explicit images, requiring time-consuming retrieval or image generation systems. 
This paper shows that
explicit images are not necessary to visually augment an LM. Instead, we use visually-grounded text representations obtained from the well-known CLIP multimodal system. For a fair comparison, we modify \textsc{VaLM}, a visually-augmented LM which uses image retrieval and representation, to work directly with visually-grounded text representations. We name this new model \textsc{Blind-VaLM}. We show that \textsc{Blind-VaLM} performs on par with \textsc{VaLM} for Visual Language Understanding (VLU), Natural Language Understanding (NLU) and Language Modeling tasks, despite being significantly more efficient and simpler. 
We also show that scaling up our model within the compute budget of \textsc{VaLM}, either increasing the model or pre-training corpus size, we outperform \textsc{VaLM} for all the evaluation tasks.  

\end{abstract}

\section{Introduction}
\label{sec:intro}

Autoregressive Language Models, such as GPT-4 \citep{achiam2023gpt} and Llama \citep{dubey2024llama}, are the reference systems for Natural Language Understanding and Generation. However, due to reporting bias in textual corpora \citep{shwartz-choi-2020-neural}, LMs lack visual knowledge, which means that they do not know the visual properties of our world, struggling to predict the typical colors, sizes and shapes of real objects, for instance \citep{alper2023:is-bert-blind, zhang2022visual, liu2022things}. Several researchers tried to overcome those problems augmenting LMs with visual knowledge \citep{tan2020vokenization, tang2021vidlankd, yang2022z, lu2022imagination}, but focusing specially on Masked Language Models (MLM). MLMs are limited for text generation and are not as versatile as autoregressive LMs. %
A recent example of visually augmenting autoregressive LMs is \textsc{VaLM} \citep{wang2022visually}, which leverages image retrieval and representation using a pretrained CLIP multimodal model \citep{radford2021learning} to improve next token prediction. To effectively use visual information, they add a Fusion Layer to a base LM, allowing textual tokens to attend visual representations before next token prediction. They show that \textsc{VaLM} improves significantly the performance for Visual Language Understanding (VLU), without degrading the NLU and text generation capabilities of the base LM.

But image retrieval and representation are very resource intensive, significantly impacting training and inference times. %
For a improved efficiency, we propose to directly use visually-grounded textual representations, obtained from the CLIP model. Based on the \textsc{VaLM} architecture, we input visually-grounded textual representations to the Fusion Layer, avoiding image retrieval and representation. We name this new model \textsc{Blind-VaLM}. As the result of our experiments we show that: i) \textsc{Blind-VaLM} is orders of magnitude faster than \textsc{VaLM} for both training and inference; ii) \textsc{Blind-VaLM} performs on par with \textsc{VaLM} for VLU, NLU and LM tasks; iii) maintaining within the compute budget of \textsc{VaLM}, but increasing the size of the pretraining corpus or the base LLM, \textsc{Blind-VaLM} improves the results of \textsc{VaLM} for all the evaluation tasks. All the code is publicly available\footnote{\url{https://github.com/paulaonta/Blind-VaLM}}. 

\begin{figure*}[t]
   \includegraphics[width=\linewidth]{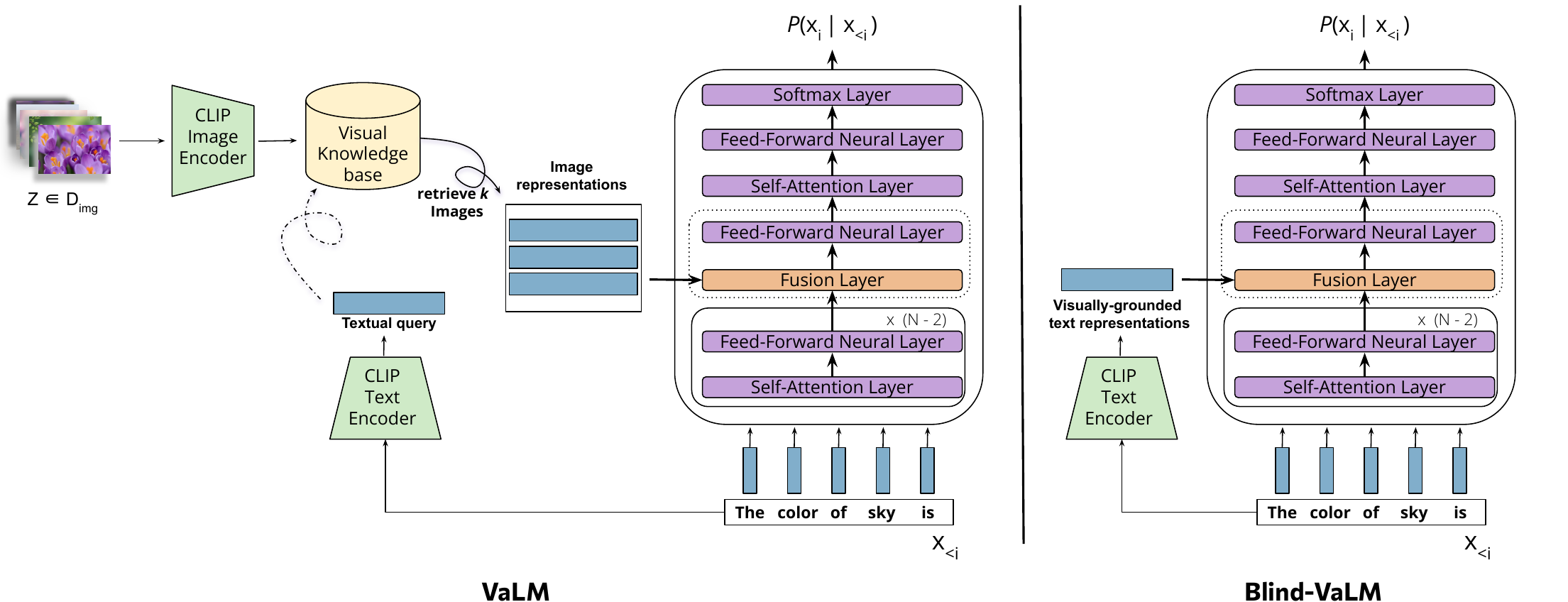}    
    \caption{Architecture comparison of the original \textsc{VaLM} (left) and our proposed \textsc{Blind-VaLM} (right).}
    \label{fig:architecture}
\end{figure*}

\section{Related Work}
\label{sec:related}

There are several approaches in the literature to augment Language Models with visual knowledge. Most of them focus on Masked Language Models (MLM) such as BERT \citep{devlin2018bert} or RoBERTa \citep{liu2019roberta}, proposing different approaches: Vokenization \citep{tan2020vokenization}, VidLanKD \citep{tang2021vidlankd}, Z-LaVI \citep{yang2022z} and iACE \citep{lu2022imagination} are good examples. Given the limitations of MLMs for text generation, \citet{tang2023learning} explore encoder-decoder architectures showing promising results for various generation tasks. But to the best of our knowledge, \textsc{VaLM} \citep{wang2022visually} is the first work for augmenting decoder-only LMs with visual knowledge. More concretely, they augment a decoder-only LM with the so-called \textit{Visual Knowledge Fusion Layer}, where contextual text representations generated by the LM are combined with the visual representations computed for retrieved images. \textsc{VaLM}, similarly to previous approaches, uses images for training and inference, adding a significant overhead to the model. We use \textsc{VaLM} as our base system, since it provides a solid framework to validate our hypothesis, i.e. LMs can be augmented without time-consuming image retrieval and representation. A similar idea is explored by concurrent work \citep{guo2023visually}, but they do not cover decoder-only LMs.

\section{\textsc{Blind-VaLM} architecture}
\label{sec:blind}

The \textsc{VaLM} architecture is composed of three main modules (Figure \ref{fig:architecture} left): 1) a backbone autoregressive LM (GPT2 \citep{radford2019language}), 2) a text-to-image retrieval module based on CLIP \citep{radford2021learning}, and 3) the Visual Knowledge Fusion Layer (Fusion Layer for short), to fuse the contextual text representations of the LM with the image representations retrieved for the input text. The intuition is that the retrieved visual representations should help to better predict the next token. For further details on the \textsc{VaLM} architecture and Fusion Layer, see \citet{wang2022visually}.

To show that image retrieval and representation are not necessary to augment the backbone LM with visual knowledge, we make one modification to the \textsc{VaLM} architecture: instead of using the CLIP image encoder representations of the retrieved images, \textsc{Blind-VaLM} directly uses CLIP text encoder representations of the text itself (see Figure \ref{fig:architecture} right).

More formally, given an input text sequence $\{ x_i \}_{i=1}^N$, let $H^0$ denote its corresponding embedding sequence and let $H^{l} = LM_l(H^{l-1}), l \in [1,L-2]$ denote the contextual representations of the backbone LM for the first $L-2$ layers. Then, let  $H_g = \textsc{CLIP}_{text}(\{ x_i \}_{i=1}^N)$ denote grounded representations over the same input text sequence obtained from CLIP. As in \textsc{VaLM}, combine both representations of the input using the Fusion Layer, such that $H^{L-1} = FusionLayer(H^{L-2}, H_g)$, and apply a final transformer layer to obtain the final representation: $H^L = LM_L(H^{L-1})$.

In other words, we remove the image retrieval aspect from \textsc{VaLM}, where $H^{L-1} = FusionLayer(H^{L-2}, \{\textsc{CLIP}_{img}(I_i)\}_{i=1}^K)$, by replacing the retrieved image representations $\textsc{CLIP}_{img}(I_i)$ with a single textual \textsc{CLIP} representation, following the intuition that the latter already encodes relevant visual information as the product of the contrastive training process with images.

\begin{table*}[t]
\caption{\textsc{Blind-VaLM} matches \textsc{VaLM} on VLU, NLU and LM tasks when trained on the same setup, while being significantly more efficient to train.}
\centering
\begin{tabular}{lccccccc} 
\toprule
\textbf{Model} & \multicolumn{3}{c}{\textbf{Color} (ACC $\uparrow$)} & \textbf{Shape} (ACC $\uparrow$) & \multicolumn{2}{c}{\textbf{Size} (ACC $\uparrow$)} & \multicolumn{1}{c}{\textbf{AVG}}  \\
 & MemoryC & ColorTerms & VCT & ShapeITC & RelativeS & TNWT & \\ \hline

 \textsc{VaLM} & 47.09 & 41.88 & 20.46 & \textbf{40.45} & \textbf{26.03} & 23.94 & 33.30 \\
 \textsc{Blind-VaLM} & \textbf{47.20} & \textbf{46.37} & \textbf{22.60} & 40.07 & 25.43 & \textbf{25.18} & \textbf{34.48} \\ 
\bottomrule
\end{tabular}

\centering

\vspace{2pt}
\begin{tabular}{lccccccc} 
\toprule
\textbf{Model} & \textbf{Wikitext-103} & \multicolumn{2}{c}{\textbf{Lambada}} & \textbf{SST-2} & \textbf{DBPedia} & \textbf{AGNews} & \textbf{MPQA} \\ 
 & PPL $\downarrow$ & PPL $\downarrow$ & ACC $\uparrow$ & ACC $\uparrow$ & ACC $\uparrow$ & ACC $\uparrow$ & ACC $\uparrow$ \\ 
 \hline

 \textsc{VaLM} & 43.68 &\textbf{ 45.69} & 35.61 &44.66& 66.77 & 41.63 & 67.95 \\
 \textsc{Blind-VaLM} & \textbf{43.46 }& 45.78 & \textbf{39.06} & \textbf{44.71} & \textbf{73.65} & \textbf{44.97} & \textbf{71.65} \\ 
\bottomrule
\end{tabular}
\label{tab:blindvsvalm}
\end{table*}

\begin{table*}[t]
\caption{When trained within the same compute budget  by increasing either model size (\textsc{Blind-VaLM-Medium}, referred to as \textsc{Blind-VaLM-M} in the table) or pre-training compute (\textsc{Blind-VaLM+}), our approach outperforms \textsc{VaLM} on both  VLU and NLU tasks.}
\centering
\begin{tabular}{lccccccc} 
\toprule
\textbf{Model} & \multicolumn{3}{c}{\textbf{Color} (ACC $\uparrow$)} & \textbf{Shape} (ACC $\uparrow$) & \multicolumn{2}{c}{\textbf{Size} (ACC $\uparrow$)} & \multicolumn{1}{c}{\textbf{AVG}}  \\
 & MemoryC & CTerms & VCT & ShapeITC & RelativeS & TNWT & \\
 \hline 
  \textsc{VaLM} & 47.09 & 41.88 & 20.46 & 40.45 & \textbf{26.03} & 23.94 & 33.30 \\
\hline 
 \textsc{Blind-VaLM} & 47.20 & 46.37 & 22.60 & 40.07 & 25.43 & 25.18 & 34.47 \\ 
 \textsc{Blind-VaLM+} &45.97  & 48.71 &20.51  & \textbf{43.64} & 25.33 &\textbf{25.40} & 34.93\\ 
 \textsc{Blind-VaLM-M} & \textbf{47.60} & \textbf{48.93} & \textbf{24.09}& 43.33  & 24.36 & 24.93 & \textbf{35.54} \\ 
\bottomrule
\end{tabular}
\vspace{3pt}
\centering
\begin{tabular}{lccccccc} 
\toprule
\textbf{Model} & \textbf{Wikitext-103} & \multicolumn{2}{c}{\textbf{Lambada}} & \textbf{SST-2} & \textbf{DBPedia} & \textbf{AGNews} & \textbf{MPQA} \\ 
 & PPL $\downarrow$ & PPL $\downarrow$ & ACC $\uparrow$ & ACC $\uparrow$ & ACC $\uparrow$ & ACC $\uparrow$ & ACC $\uparrow$ \\ 
 \hline 
  \textsc{VaLM} & 43.68 & 45.69 & 35.61 &44.66& 66.77 & 41.63 & 67.95 \\
  \hline 
 \textsc{Blind-VaLM} & 43.46 & 45.78 & 39.06 & 44.71 & 73.65 & 44.97 & 71.65 \\ 
 \textsc{Blind-VaLM+} & 40.95 & 43.83  &39.05  & 44.66  & \textbf{74.42} & \textbf{50.44}&  \textbf{78.10}\\ 
 \textsc{Blind-VaLM-M} &\textbf{33.95 }&  \textbf{37.63} & \textbf{45.08}   & \textbf{59.00} & 70.57 & 48.28 & 55.65  \\
\bottomrule
\end{tabular}
\label{tab:scaling}
\end{table*}

\section{Experimental setup}
\label{sec:experiments}
We want to show that actual retrieval and representation of images is not necessary to augment a LM with visual knowledge, and that directly using textual representations from a visually grounded text encoder instead is sufficient. To that end, we initially pre-train \textsc{Blind-VaLM} and \textsc{VaLM} models in a comparable setting, which we now describe.

\paragraph{Text Corpus.} Following the original \textsc{VaLM} \citep{wang2022visually}, we use the English corpus of CC-100 \citep{conneau2020unsupervised} as the pre-training text corpus for all models. 
Due to limited access to compute, we only consume 10.5B tokens for  pre-training, which accounts for around 19\% of the English portion of the corpus.

\paragraph{Image data and retrieval module.} \textsc{VaLM} requires an image database and a vector retrieval module. We use a \emph{FAISS} \citep{johnson2019billion} index, trained on the exact same setup as the original \textsc{VaLM}, based on the \textsc{GPT2-Small} architecture, as detailed in  Appendix \ref{app:hparams}.%

\paragraph{Pre-training hyperparameters.} We train both models on the exact same setup, following a configuration similar to the original \textsc{VaLM}. Details can be found in Appendix \ref{app:hparams}.

Due to the increased efficiency of our architecture, our model employs significantly less compute than the original: \textsc{Blind-VaLM} was trained on 530 GPU-hours, while the \textsc{VaLM} baseline required 1.2K GPU-hours, making our approach \textbf{2.2x faster to train}. We train all our models on a cluster of 8 A100 GPUs.

\paragraph{Evaluation.} We evaluate our models on VLU, NLU and LM tasks (see Appendix \ref{app:prompts} for details).

For \textbf{VLU}, we focus on three basic visual properties of objects: color, shape and size. We evaluate color knowlege on the following datasets: Memory Color \citep{norlund2021transferring}, Color Terms \citep{bruni2012distributional} and ViComTe (color subset) \citep{zhang2022visual}. We evaluate knowledge about shape on the ShapeITC \citep{alper2023:is-bert-blind} dataset, and knowledge of size on the RelativeSize \citep{bagherinezhad2016elephants} and Things Not Written in Text \citep{liu2022things} datasets.

We evaluate \textbf{NLU} capabilities on four downstream tasks: two sentiment analysis tasks on the SST-2 and MPQA datasets \citep{socher2013recursive,wiebe2005annotating}, and two topic classification tasks on the AGNews and DBPedia datasets \citep{auer2007dbpedia, zhang2015character}. Additionally, we evaluate pure language modeling ability by measuring perplexity on the Wikitext-103 and Lambada datasets \citep{merity2016pointer, paperno2016lambada}. For Lambada, we also report accuracy at predicting the last word of each sentence, following the original \textsc{VaLM} work.

\paragraph{Scaling \textsc{Blind-VaLM}.} It is important to note that, since \textsc{Blind-VaLM} does not require an actual image retrieval step, it is significantly more efficient at both training and inference time.\footnote{The efficiency gains are two-fold: 1) since we only use a single CLIP text representation in the Fusion Layer, this layer requires less floating-point operations (FLOPs), and 2) since we do away with the dense vector database retrieval, training and inference latency is significantly reduced, leading to improved wall-clock efficiency.} Taking advantage of this increased efficiency, we explore two ways of scaling up \textsc{Blind-VaLM}, within the compute budget of our \textsc{VaLM} baseline.

\paragraph{(i) Scaling model size.} We train \textbf{\textsc{Blind-VaLM-Medium}}, by switching the LM backbone architecture to \textsc{GPT2-Medium} instead of  \textsc{GPT2-Small}. See Appendix \ref{app:hparams} for details. This larger model was trained on 595 GPU-hours, still within the compute budget of the \textsc{VaLM} baseline.

\paragraph{(ii) Scaling pre-training compute.} We train \textbf{\textsc{Blind-VaLM+}}, by simply continuing pre-training the baseline \textsc{Blind-VaLM} for longer, until we reach a total of 88255 steps, which corresponds to 23.1B tokens (42\% of CC-100). Again, thanks to the increased efficiency of our approach, this model is compute-matched\footnote{Note that we use \textit{compute-matched} in the wall-clock time sense here, not the  FLOPs sense, since the time consuming dense vector retrieval step we remove is not reflected by FLOPs.} to the original \textsc{VaLM} baseline, taking a total of 1.17K GPU-hours to train.

\section{Results}

We next describe our main results.

\paragraph{\textsc{Blind-VaLM} matches \textsc{VaLM} on VLU, NLU and LM tasks.} Table \ref{tab:blindvsvalm} shows results for \textsc{Blind-VaLM} and \textsc{VaLM} trained on the same setup, as described in \S\ref{sec:experiments}. We observe that our approach matches the original \textsc{VaLM} on VLU tasks, as well as NLU and LM tasks: it achieves an average score $1.18$ points higher in VLU tasks, and outperforms \textsc{VaLM} on 6/7 NLU \& LM tasks. This supports our hypothesis that \textbf{actually retrieving and encoding images is not required for visual augmentation}, since simply utilizing textual representations from an already visually grounded text encoder works equally well. Additionally, as described in section \S\ref{sec:experiments} \textbf{\textsc{Blind-VaLM} is 2.2x faster} to train, since it skips the time consuming vector retrieval step. Speedups are even more significant at inference time, since generation is not as compute-bound and retrieval latency plays a bigger role.

\paragraph{\textsc{Blind-VaLM} outperforms \textsc{VaLM}, when trained within the same  compute budget.} Table \ref{tab:scaling} shows results for two scaled-up \textsc{Blind-VaLM} variants, obtained through scaling either pre-training compute or model size, as described in \S\ref{sec:experiments}. We observe that both variants outperform \textsc{VaLM}, while being trained within the same compute budget as it. For example, in the case of \textsc{Blind-VaLM-Medium}, we outperform \textsc{VaLM} by $2.2$ points on average for VLU tasks, and outperform \textsc{VaLM} on 6/7 NLU \& LM tasks.

\section{Conclusions}
\label{sec:conclusions}

In this work we test the hypothesis that explicit image retrieval is not necessary to augment an LM with visual information. To that end, we train a modified variant of \textsc{VaLM} \citep{wang2022visually}, which we call \textsc{Blind-VaLM}, by replacing the retrieved image encoding vectors with textual embeddings obtained from the visually grounded CLIP encoder \citep{radford2021learning}.

Our results show that \textsc{Blind-VaLM} matches \textsc{VaLM} when trained on the same data, while being significantly more efficient to train. Additionally, scaling up our model within the compute budget of \textsc{VaLM},
our approach outperforms \textsc{VaLM}. Overall, these results show that simply leveraging the textual encoding from an already visually grounded CLIP encoder is enough to obtain the same gains on visual tasks as \textsc{VaLM}, supporting our hypothesis that actual image retrieval is not essential.

These results open up new avenues in the line of work of visually augmenting language models, beyond the paradigm of image retrieval. 
Our findings allow for more efficient visual augmentations, which will result on broader exploration capacity for future works.

\section{Limitations}
\label{sec:limitations}
Our work is focused on English only, due to the size and accessibility of the resources, i.e. text corpora, image-text datasets, different pre-trained models and evaluation benchmarks. However, it would be very interesting to extend the work to other languages.

The VLU evaluation is limited to visual object properties, such as color, shape and size. But, visual language is broader and extending evaluation benchmarks would allow to better understand how VLU evolves in pure and visually-augmented LMs.

Finally, we use the original CLIP model \citep{radford2021learning} to visually-augment LMs, as done in \textsc{VaLM}. Nowadays there are more powerful multimodal models and it would be very interesting to explore how those better models impact in the knowledge acquisition of visually-augmented LMs.

\section*{Acknowledgments}
This work is partially supported by the Ministry of Science and Innovation of the Spanish Government (AWARE project TED2021-131617B-I00, DeepR3 project TED2021-130295B-C31), project funded by MCIN/AEI/10.13039/501100011033 and European Union NextGeneration EU/PRTR, and the Basque Government (IXA excellence research group IT1570-22 and IKER-GAITU project).

\bibliography{mybibfile}

\appendix

\section{Training hyper-parameters}
\label{app:hparams}
We train all models using an inverse-square-root learning rate schedule, the \textsc{Adam} optimizer \citep{kingma2014adam} with $\beta_1 = 0.9$, $\beta_2 = 0.98$, a peak LR of $2e-3$, a weight-decay of $0.01$, dropout of $0.1$, $4000$ warmup steps, a global batch size of $512$, and a sequence length of $512$ tokens. We train both models for 40600 steps.

Since both \textsc{Blind-VaLM} and the original \textsc{VaLM} share the limitation of needing tokenization to be compatible with the CLIP used for retrieval, use use the same BPE tokenizer as \textsc{CLIP} and the original \textsc{GPT2}.

\textsc{VaLM}, \textsc{Blind-VaLM} and \textsc{Blind-VaLM}+ use the \textsc{GPT2-Small} architecture, comprising 124M parameters, for the LM backbone, and the \textsc{CLIP-RN50x16 } model for the visually grounded text-image encoder, with a total of 85M parameters on the text encoder.

For \textsc{Blind-VaLM-Medium}, we switch the LM backbone architecture to \textsc{GPT2-Medium}, with a total of 345M parameters. We switch the CLIP encoder to the \textsc{CLIP-RN50x64 } version with 151M parameters, which employs the hidden-size corresponding to \textsc{GPT2-Medium}, required by the Fusion Layer. 

\section{Evaluation details}
\label{app:prompts}
To evaluate VLU and NLU capabilities we follow the prompting approach designed by \textsc{VaLM} \citep{wang2022visually}, with some slight modifications. Tables \ref{table:prompt_color}, \ref{table:prompt_shape} and \ref{table:prompt_size} show the prompts we use for color, shape and size evaluation. Similarly, Table \ref{table:prompt_nlu} defines the prompts for all the NLU tasks we consider. The reported results in the paper are always the average of the results obtained for all the prompts for a given task, with the objective of avoiding any bias towards different prompting choices.

\begin{table*}[htbp]                                             
    \caption{  The prompts and prediction labels used in object color reasoning.}          
    \centering
    \small
    \setlength{\tabcolsep}{1.0mm}{
    \scalebox{0.9}{
        \begin{tabular}{ p{2cm} | p{2.5cm}  p{9cm} l}
            \hline
            \toprule
            \textbf{Task} &  \textbf{Dataset}&  \textbf{Prompt} & \textbf{Labels} \\
            \midrule
            \multirow{17}{2cm}{\textbf{Object Color Reasoning}} & \multirow{9}{2.3cm}{\centering Memory Colors and Color Terms} &Q: What is the color of [DESCRIPTOR] [ITEM]? A: It is [Label] & \multirow{9}{3cm}{\{red, white, orange, green, blue, yellow, purple, black, pink, grey, brown\}} \\
            && Q: What is the colour of [DESCRIPTOR] [ITEM] ? A: It is [Label] \\
            && What is the color of [DESCRIPTOR] [ITEM]? It is [Label] \\
            & &What is the colour of [DESCRIPTOR] [ITEM]? [Label] \\
             & &The color of [DESCRIPTOR] [ITEM] is [Label] \\
            & &The usual color of [DESCRIPTOR] [ITEM] is [Label] \\
            & &[DESCRIPTOR] [ITEM] usually has the color of [Label] \\
            & &What is the usual color of [DESCRIPTOR] [ITEM]? [Label] \\
            & &What is the typical color of [DESCRIPTOR] [ITEM]? [Label] \\
            &&&\\
            & \multirow{7}{2cm}{\centering ViComTe}&[ITEM] can be of color & \multirow{7}{3cm}{\{red, white, orange, green, blue, yellow, purple, black, pink, grey, brown, silver\}} \\
            && [ITEM] has color\\
            && The color of [ITEM] can be \\
            & &The color of the [ITEM] is\\
            & &[ITEM] is \\
            & &This [ITEM] is  \\
            & &[ITEM] is of color \\
            \bottomrule
            \hline
            \end{tabular}
        }
    }
    \label{table:prompt_color}
\end{table*}

\begin{table*}[htbp]                                             
    \caption{ The prompts and prediction labels used in object shape reasoning.}                   
    \centering
    \small
    \setlength{\tabcolsep}{1.0mm}{
    \scalebox{0.9}{
        \begin{tabular}{ p{2cm} | p{5cm} l}
            \hline
            \toprule
            \textbf{Task} & \textbf{Prompt} & \textbf{Labels} \\
            \midrule
            \multirow{9}{2cm}{\textbf{Object Shape Reasoning}} &  [ITEM] can be shape of & \multirow{9}{2.5cm}{\{circle, rectangle, triangle\}} \\
            &[ITEM] has shape of\\
            &[ITEM] is of shape\\
            &The shape of [ITEM] can be\\
            &The shape of the [ITEM] is\\
            &[ITEM] is\\
            &This [ITEM] is\\
            &[ITEM] can be shape\\
            &[ITEM] has shape\\
            \bottomrule
            \hline
            \end{tabular}
        }
    }
    \label{table:prompt_shape}
\end{table*}

\begin{table*}[htbp]  
    \caption{ The prompts and prediction labels used in object size reasoning.}      
    \centering
    \small
    \setlength{\tabcolsep}{1.0mm}{
    \scalebox{0.9}{
        \begin{tabular}{ p{2cm} | p{2.5cm}  p{8.5cm} l}
            \hline
            \toprule
            \textbf{Task} &  \textbf{}& \textbf{Prompt} & \textbf{Labels} \\
            \midrule
            \multirow{16}{2cm}{\textbf{Object Size Reasoning}} & \multirow{10}{2.3cm}{\centering  Object prediction} &Which is bigger? [ITEMA] or [ITEMB] & \multirow{10}{2cm}{\{ITEMA, ITEMB\}} \\
            &&Which is smaller? [ITEMA] or [ITEMB]\\
            &&Which is larger? [ITEMA] or [ITEMB]\\
            &&Which is tinier? [ITEMA] or [ITEMB]\\
            &&Which has a bigger size? [ITEMA] or [ITEMB]\\
            &&Which has a bigger size? [ITEMA] or [ITEMB]\\
            &&Which has a smaller size? [ITEMA] or [ITEMB]\\
            &&Which one is larger in size? [ITEMA] or [ITEMB]\\
            &&Which one is smaller in size? [ITEMA] or [ITEMB]\\
            &&&\\
            &\multirow{6}{2.3cm}{\centering Size prediction}&[ITEMA] is larger or smaller than [ITEMB]? & \multirow{6}{2cm}{\{larger, smaller\}}\\
            &&[ITEMB] is larger or smaller than [ITEMA]?\\
            &&The size of [ITEMA] is larger or smaller than [ITEMB]?\\
            &&The size of [ITEMB] is larger or smaller than [ITEMA]?\\
            &&[ITEMA] has a larger or smaller size than [ITEMB]?\\
            &&[ITEMB] has a larger or a smaller size than [ITEMA]?\\        
            \bottomrule
            \hline
            \end{tabular}
        }
    }                                            
    \label{table:prompt_size}
\end{table*}

\begin{table*}[htbp]  
    \caption{ The prompts and prediction labels used in 4 natural language understanding datasets.
The labels for AGNews are \{world, sports, business, technology\} and the labels for DBPedia are \{company, school, artist, athlete, politics, transportation, building, nature, village, animal, plant, album, film, book\}.}                   
    \centering
    \small
    \setlength{\tabcolsep}{1.0mm}{
    \scalebox{0.9}{
        \begin{tabular}{ p{2.5cm} | p{2cm}  p{6cm} l}
            \hline
            \toprule
            \textbf{Task} &  \textbf{Dataset}& \textbf{Prompt} & \textbf{Labels} \\
            \midrule
            \multirow{4}{2.5cm}{\textbf{Natural Language Understanding}} &  SST-2 &Review: [Sentence] Sentiment: [Label] & \{Positive, Negative\} \\
            & MPQA&Review: [Sentence] Sentiment: [Label]&\{Positive, Negative\} \\
            &AGNews&input: [Sentence] type: [Label]& \{world,\dots, technology\} \\
            &DBPedia&input: [Sentence] type: [Label]&  \{company, school,\dots, book\}\\       
            \bottomrule
            \hline
            \end{tabular}
        }
    }           
    \label{table:prompt_nlu}
\end{table*}

\end{document}